%% file: root.tex
\documentclass[letterpaper, 10 pt, journal, twoside]{IEEEtran}

\usepackage{amsmath,amsfonts}
\usepackage{array}
\usepackage[caption=false,font=normalsize,labelfont=sf,textfont=sf]{subfig}
\usepackage{textcomp}
\usepackage{stfloats}
\usepackage{url}
\usepackage{verbatim}
\usepackage{graphicx}
\usepackage{cite}

\usepackage{times}
\usepackage[dvipsnames]{xcolor}
\usepackage{wrapfig}
\usepackage{ulem}
\usepackage{amssymb}
\usepackage{booktabs}
\usepackage{silence}
\usepackage{utfsym}
\usepackage{dsfont}
\usepackage{mathabx}
\usepackage{tabularx}
\usepackage{float}
\usepackage{colortbl}
\usepackage{enumerate}
\usepackage{multicol,multirow}
\usepackage{setspace}
\usepackage{makecell}
\usepackage{pifont}
\usepackage{framed}
\usepackage{titletoc}
\usepackage{overpic}
\usepackage{textpos}
\usepackage{titlesec}
\usepackage{comment}
\usepackage{listings}
\usepackage{bbm}
\usepackage[linesnumbered,ruled,vlined]{algorithm2e}
\usepackage[colorlinks=false,hidelinks]{hyperref}

\DeclareMathOperator*{\argmax}{arg\,max}

\definecolor{myblue}{RGB}{43, 99, 179}
\definecolor{myred}{RGB}{165,34,21}
\definecolor{mygrey}{RGB}{150,150,150}
\definecolor{DeepBlue}{RGB}{65,100,170}

\title{ARMADA: Autonomous Online Failure Detection and Human Shared Control Empower Scalable Real-world Deployment and Adaptation}

\author{Wenye Yu$^{1,2}$, Jun Lv$^{1,3}$, Zixi Ying$^{3}$, Yang Jin$^{1}$, Chuan Wen$^{1,\dagger}$, and Cewu Lu$^{1,2,3,\dagger}$%
\thanks{Manuscript received: February 7, 2026; Revised May 4, 2026; Accepted June 24, 2026.}%
\thanks{This paper was recommended for publication by Editor Editor A. Name upon evaluation of the Associate Editor and Reviewers' comments.
This work was supported by organizations/grants which supported the work.}%
\thanks{$^{1}$Wenye Yu, Jun Lv, Yang Jin, Chuan Wen, and Cewu Lu are with Shanghai Jiao Tong University.}%
\thanks{$^{2}$Wenye Yu and Cewu Lu are with Shanghai Innovation Institute.}%
\thanks{$^{3}$Jun Lv, Zixi Ying, and Cewu Lu are with Noematrix Ltd.}%
\thanks{$^{\dagger}$Chuan Wen and Cewu Lu are equal advisors. Project page: \url{https://virlus.github.io/armada/}.}%
\thanks{Digital Object Identifier (DOI): see top of this page.}%
}

\begin{document}

\maketitle

\begin{abstract}
\input{tex/0_abstract}
\end{abstract}


\input{tex/1_intro}

\begin{figure*}[!t]
    \centering
    \includegraphics[width=0.7\linewidth]{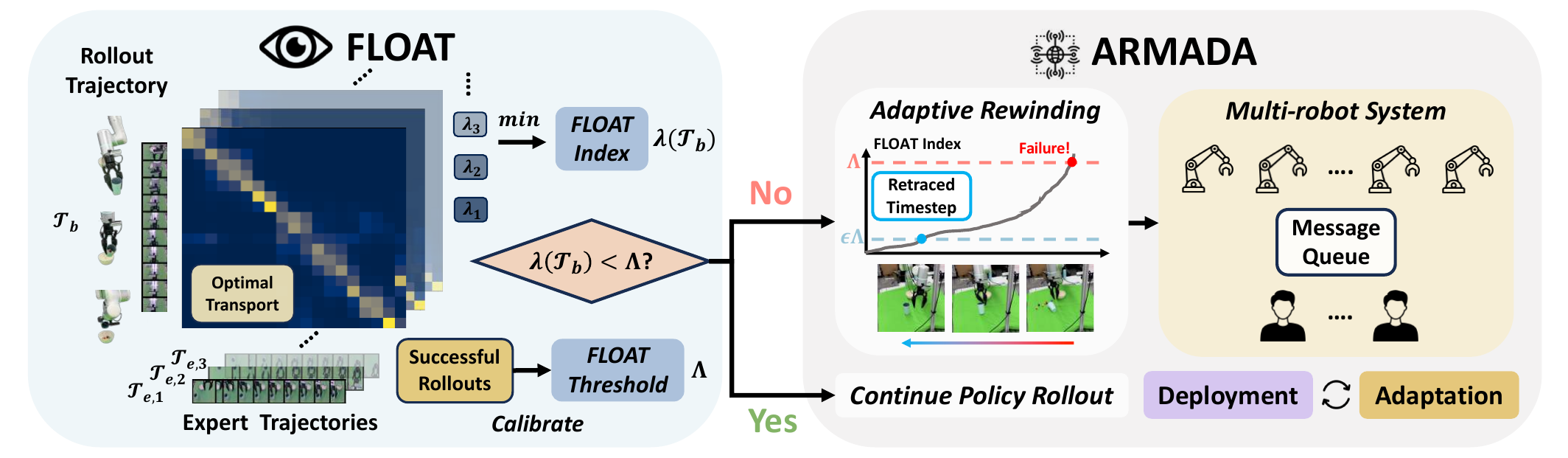}
    \caption{\textbf{Method overview.}
    FLOAT failure detector conducts real-time OT matching between the policy embeddings of the current rollout and all expert demonstrations, and defines the minimum OT cost as FLOAT index.
    We thereby calibrate the FLOAT threshold on all successful rollouts.
    When the FLOAT index of a rollout trajectory exceeds the threshold, we consider it a failure and employ adaptive rewinding based on OT computation, which helps retrace a previous timestep before the scene was disturbed.
    Our multi-robot system then allocates an idle human operator to the failed robot for intervention.
    }
    \label{fig:method}
\end{figure*}

\input{tex/2_related_work}

\input{tex/3_preliminaries}

\input{tex/4_method}

\input{tex/5_experiment}

\input{tex/6_discussion}

\input{tex/7_conclusion}

{\tiny
\bibliographystyle{IEEEtran}
\bibliography{references}
}

\end{document}

%% file: tex/0_abstract.tex
Imitation learning has shown promise in learning from large-scale real-world datasets.
However, pretrained policies usually perform poorly without sufficient in-domain data.
Besides, human-collected demonstrations entail substantial labour and tend to encompass mixed-quality data and redundant information.
As a workaround, human-in-the-loop systems gather domain-specific data for policy post-training, and exploit closed-loop policy feedback to offer informative guidance, but usually require full-time human surveillance during policy rollout.
In this work, we devise ARMADA, a multi-robot deployment and adaptation system with human-in-the-loop shared control, featuring an autonomous online failure detection method named FLOAT.
Thanks to FLOAT, ARMADA enables paralleled policy rollout and requests human intervention only when necessary, significantly reducing reliance on human supervision.
Hence, ARMADA enables efficient acquisition of in-domain data, and leads to more scalable deployment and faster adaptation to new scenarios.
We evaluate the performance of ARMADA on four real-world tasks.
FLOAT achieves nearly 95\% accuracy on average, surpassing prior state-of-the-art failure detection approaches by over 20\%.
Besides, ARMADA manifests more than 4$\times$ increase in success rate and greater than 2$\times$ reduction in human intervention rate over multiple rounds of policy rollout and post-training, compared to previous human-in-the-loop learning methods.

%% file: tex/1_intro.tex
\section{Introduction}

\IEEEPARstart{R}{ecent} years have seen rapid growth in data-driven robotic manipulation~\cite{chi2023diffusion,zhao2023learning,kim2024openvla}.
To support real-world deployment, recent efforts have focused on building large-scale datasets for policy learning.
Some works aggregate human-collected data from multiple sources into large heterogeneous datasets~\cite{o2024open,khazatsky2024droid,fang2024rh20t}. 

Another line of work develops unified data collection systems, including embodiment-agnostic handheld devices \cite{chi2024universal,tao2025dexwild, xu2025dexumi} and low-cost exoskeletons \cite{fang2024airexo,ben2025homie,fang2025airexo}.
Pretraining on these datasets yields policies with rudimentary real-world task capabilities.

However, these pretrained policies often lack robustness at deployment because in-domain data are scarce.
Moreover, as noted in prior work on data curation \cite{hejna2025robot,chen2025curating,zhang2025scizor}, human demonstrations often contain mixed-quality and redundant segments that hinder performance.
\begin{figure}[!t]
    \vspace{-3mm}
    \centering
    \includegraphics[width=0.95\linewidth]{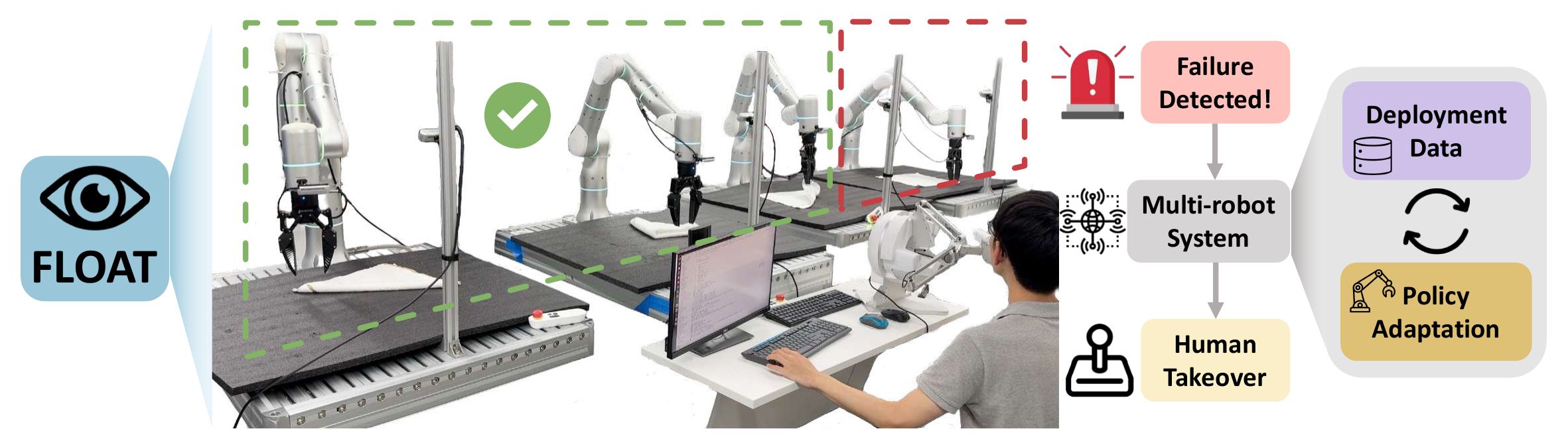}
    \caption{\textbf{Illustration of ARMADA.} 
    ARMADA uses the FLOAT failure detector to enable parallel policy rollouts on multiple robots, requesting human intervention only when needed.
    The resulting online deployment data are then used for policy improvement, forming a scalable deployment-adaptation loop.}
    \label{fig:teaser}
    \vspace{-4mm}
\end{figure}

Prior works therefore explore human-in-the-loop systems, where human operators and learned policies share robot control.
These systems enable interactive collection of domain-specific data for post-training and adaptation.
By observing online rollouts, human operators can also provide informative closed-loop feedback.
Some methods let humans intervene when the robot fails and use the correction data for post-training \cite{liu2022robot,liu2024multi,wang2025genie,kasaei2024vital,wu2025robocopilot}.
Others pursue shared autonomy through time-varying control ratios between human and robot \cite{luo2024human,yoneda2023noise}.

Nevertheless, most human-in-the-loop systems still require full-time human supervision so operators can detect failures and assist recovery.
As a result, they remain largely limited to one operator per robot.
To improve scalability, we identify two desiderata: 
\textbf{1)} A real-time failure detector that monitors policy rollouts and alerts operators when failures are likely, reducing the need for constant supervision.
\textbf{2)} A multi-robot shared-control system that scales deployment and post-training for faster adaptation.

We therefore propose \textbf{ARMADA} (\textbf{A}utonomous \textbf{R}eal-world \textbf{M}ulti-robot system with human \textbf{A}ssistance for \textbf{D}eployment and \textbf{A}daptation), a scalable real-world robot system with autonomous online failure detection and multi-robot shared control.
Concretely, we devise an online failure detector for visuomotor imitation learning based on policy embeddings.
Given a pretrained visuomotor policy and its training demonstrations, we match online rollouts to expert trajectories and compute their embedding distance via Optimal Transport (OT) \cite{haldar2023watch,papagiannis2022imitation,dadashi2020primal}, which we call the \textbf{FLOAT} index (\textbf{F}ai\textbf{L}ure detection based on \textbf{O}ptim\textbf{A}l \textbf{T}ransport).
Using FLOAT indices from successful rollouts, we define a universal threshold for real-time, plug-and-play failure detection.
We further implement a multi-robot shared-control system that supports autonomous rollouts and timely human intervention when FLOAT raises a warning, substantially improving efficiency.
At a high level, ARMADA forms a closed loop: the policy roll outs online, FLOAT monitors for failure, adaptive rewinding returns the robot to a recoverable state when needed, a human then intervenes, and the rollout continues for data collection and post-training.
Our system readily adapts to different imitation learning policies, intervention interfaces (teleoperation, exoskeleton, etc.), and embodiments.
The entire implementation of ARMADA will be open-sourced.

We conduct comprehensive experiments to evaluate FLOAT and ARMADA.
Across four real-world tasks, FLOAT raises average accuracy to nearly 95\%, over 20\% higher than state-of-the-art baselines.
Over several rounds of rollout and fine-tuning, ARMADA achieves more than a 4$\times$ increase in success rate and more than a two-fold reduction in human intervention rate over prior human-in-the-loop shared-control systems.

In a nutshell, our contributions are three-fold:

\begin{enumerate}
    \item We devise FLOAT, a plug-and-play online failure detector for visuomotor imitation learning that achieves nearly 95\% accuracy in the real world.
    \item We implement ARMADA, a distributed multi-robot human-in-the-loop shared-control system that enables parallel autonomous rollouts for scalable real-world deployment and adaptation.
    \item Over multiple rounds of post-training, ARMADA delivers more than a four-fold increase in success rate and more than a two-fold reduction in human intervention ratio over prior human-in-the-loop methods that require full-time supervision.
\end{enumerate}

%% file: tex/2_related_work.tex
\section{Related work}

\subsection{Human-in-the-loop learning in robot manipulation}

Human-in-the-loop learning leverages interactive human signals to guide policy learning \cite{kasaei2024vital, mandlekar2023human, torne2023breadcrumbs}.
By incorporating human prior knowledge, these methods introduce strong inductive bias.
Human input can take several forms.
Some works use preference rankings to align robot behaviour with human preferences via reinforcement learning \cite{tian2024maximizing,hejna2023few,kuhar2023learning, chen2025fdpp, zhang2024grape}, while others steer policies at inference time without fine-tuning \cite{nakamoto2024steering, wu2025foresight, wang2024inference}.
Shared-autonomy methods let humans and policies share control with a time-varying ratio \cite{luo2024human,yoneda2023noise}.
Other methods request human intervention only when failure is likely and use the resulting corrections to improve policies \cite{liu2022robot,wu2025robocopilot, luo2024precise, liu2024multi}.
Our system belongs to the last category, but our online failure detector removes the need for continuous monitoring and enables one-to-multiple shared control, accelerating real-world deployment and adaptation.

\subsection{Online failure detection for pretrained policies}

Failure detection is critical when deploying pretrained policies, especially imitation-learned ones that are vulnerable to out-of-distribution environments \cite{yang2025novel, sinha2022system, lu2025robofac}.
Many methods cast this problem as OOD detection.
For example, prior work trains failure classifiers from rollout data or detects OOD cases by K-Means clustering on policy embeddings \cite{liu2024model,liu2024multi}, but such methods require pre-collected failure data and scale poorly in real-world settings.

Other approaches use VAE reconstruction error \cite{wong2022error}, time-varying conformal prediction bands \cite{xu2025can}, or diffusion-based policy's uncertainty \cite{chen2025uncertainty, he2025uncertainty, lee2025diff}.
However, these methods are sensitive to environmental changes and prone to overfitting, as we show in our experiments.
Some works instead use Large Language Models (LLMs) or Vision-Language Models (VLMs): Sinha et al.~\cite{sinha2024real} build a classifier from LLM embeddings, while Sentinel \cite{agia2024unpacking} and Genie Centurion \cite{wang2025genie} directly query VLMs for failure judgments.
These methods either rely on zero-shot generalization of VLMs or require domain-specific fine-tuning.
Agia et al.~\cite{agia2024unpacking} propose statistical temporal action consistency (STAC), the current state-of-the-art failure detection approach.
Our method provides plug-in-and-play failure detection for visuomotor imitation learning, using only policy embeddings and improving accuracy by over 20\% relative to STAC, as detailed in Sec.~\ref{sec:exp}.

\subsection{Post-training of imitation learning policies}

Data quality strongly affects policy post-training \cite{belkhale2023data,lin2024data,mandlekar2021matters,saxena2025matters}.
However, human-collected demonstrations often contain mixed-quality and redundant data \cite{zhang2025scizor,hejna2025robot}.
To filter undesirable trajectories, prior work proposes several data-curation strategies.
SCIZOR \cite{zhang2025scizor} removes redundant and suboptimal trajectories in a self-supervised manner, and DemInf \cite{hejna2025robot} estimates data quality via mutual information.
Other methods estimate each demonstration's influence through online rollouts and refine the training set accordingly \cite{agia2025cupid,chen2025curating}.
Hejna et al.~\cite{hejna2024re} optimize dataset mixtures automatically for large heterogeneous datasets, while Octo \cite{team2024octo} uses heuristic dataset weighting.
We instead improve data quality with an adaptive rewinding mechanism: by returning to an earlier timestep, the robot can recover from failure while a human resets the scene, yielding informative corrective demonstrations.

%% file: tex/3_preliminaries.tex
\section{Preliminaries}

\subsection{Problem formulation}
\label{subsec:problem_formulation}

We consider real-world deployment of a pretrained policy under an iterative post-training setting. In each rollout episode $\mathcal{T}_b=\{(o_{b,t},a_{b,t})\}_{t=1}^{l_b}$, an online anomaly detector monitors the partial trajectory $(\mathcal{T}_b)_{1:t_0}$ and judges whether the current episode has already failed before termination. Once failure is detected, human intervention can be requested to correct the rollout and avoid uninformative data collection.

Formally, let $\mathcal{D}_e$ denote the offline expert dataset and $\pi_0$ the initial imitation-learned policy. At deployment round $m$, policy $\pi_m$ interacts with the environment to collect rollout and intervention data $\mathcal{D}_{b,m}$. The aggregated training set is then updated as $\mathcal{D}_m=\mathcal{D}_e\cup\bigcup_{k=0}^{m}\mathcal{D}_{b,k}$, and the policy is post-trained on $\mathcal{D}_m$ to obtain $\pi_{m+1}$. Repeating this DAgger-style procedure yields a sequence of progressively improved policies $\pi_0,\pi_1,\ldots,\pi_M$. Our objective is therefore twofold: to detect task failure online during policy execution, and to iteratively improve the policy through multi-round rollout and post-training.

\subsection{Optimal Transport and its application in robotics}
\label{subsec:OT}

Optimal Transport (OT) studies how to move mass between distributions with minimal cost.
In imitation learning, the goal is to optimize a behaviour policy $\pi_b$ to stay close to the expert policy $\pi_e$, given offline expert demonstrations $\mathcal{D}_e=\{\{(o_{e,t}^n,a_{e,t}^n)\}_{t=1}^{l_n}\}_{n=1}^{N}$.
Here, $N$ is the number of expert trajectories and $l_n$ is the length of the $n$-th demonstration.
OT thus provides a non-parametric way to match trajectories and compute rewards for both imitation learning \cite{papagiannis2022imitation,ng2000algorithms,abbeel2004apprenticeship} and reinforcement learning \cite{haldar2023watch,dadashi2020primal}.
Specifically, given an expert demonstration $\mathcal{T}_e=\{(o_{e,t},a_{e,t})\}_{t=1}^{l_e}$ and a rollout trajectory $\mathcal{T}_b=\{(o_{b,t}, a_{b,t})\}_{t=1}^{l_b}$, OT computes an optimal transport matrix $(\mu_{ij}^{\ast})^{l_e \times l_b}$ using the objective in Eq.~\ref{eq:single_obj}, where $c(\cdot,\cdot)$ is the cost function.
For simplicity, we write $\mu^*=\mathbf{OT}(\mathcal{T}_e,\mathcal{T}_b)$.

\begin{equation}
    \min\limits_{\mu_{ij}\geq 0} \sum\limits_{i=1}^{l_e}\sum\limits_{j=1}^{l_b}c(o_{e,i},o_{b,j}) \mu_{ij}\ \ \ s.t. \sum\limits_{j=1}^{l_b}\mu_{ij} = \frac{1}{l_e}, \sum\limits_{i=1}^{l_e}\mu_{ij} = \frac{1}{l_b}
    \label{eq:single_obj}
\end{equation}

The reward at each rollout timestep is computed by Eq.~\ref{eq:reward_compute}.

\begin{equation}
    r_t = -\sum\limits_{i=1}^{l_e} c(o_{e,i},o_{b,t}) \mu_{i,t}^{\ast}
    \label{eq:reward_compute}
\end{equation}

In this work, rather than using OT for reward computation, we propose an OT-based failure detection index, FLOAT, detailed in Sec.~\ref{sec:method}.

%% file: tex/4_method.tex
\section{Method}
\label{sec:method}

For scalable real-world deployment and adaptation, we propose ARMADA, a multi-robot shared control system equipped with FLOAT, an online failure detector, and an adaptive rewinding mechanism.
This section introduces FLOAT in Sec.~\ref{subsec:fail_detect}, details ARMADA in Sec.~\ref{subsec:multi-robot-sys}, and presents our policy architecture and training recipe in Sec.~\ref{subsec:policy_learning}.
Figure~\ref{fig:method} provides an overview of the method.

\subsection{FLOAT: Our failure detector}
\label{subsec:fail_detect}

We propose FLOAT, a failure detector for visuomotor imitation learning.
Let a policy $\pi$ with observation encoder $\phi$ be trained on expert demonstrations $\mathcal{D}_e=\{\mathcal{T}_{e,n}\}_{n=1}^{N}$, where $\mathcal{T}_{e,n}=\{(o_{e,t}^n,a_{e,t}^n)\}_{t=1}^{l_n}$.
We first extract policy embeddings for all expert demonstrations, as shown in \ref{eq:expert_embedding}.

\begin{equation}
    \mathcal{F}_e=\{\{\phi(o_{e,t}^n)\}_{t=1}^{l_n}\}_{n=1}^{N}
    \label{eq:expert_embedding}
\end{equation}

During rollout, we similarly extract policy embeddings up to the current step $t_0$, as shown in \ref{eq:rollout_embedding}.

\begin{equation}
    \mathcal{F}_b=\{\phi(o_{b,t})\}_{t=1}^{t_0}
    \label{eq:rollout_embedding}
\end{equation}

Following Sec.~\ref{subsec:OT}, we use cosine similarity as the OT cost:
$c(o_{e,i}, o_{b,j}):=\cos(\phi(o_{e,i}),\phi(o_{b,j}))$.
We then compute the OT plan between the current rollout trajectory $\mathcal{T}_b=\{(o_{b,t}, a_{b,t})\}_{t=1}^{t_0}$ and each expert demonstration $\mathcal{T}_{e,n}$, denoted by $\mu_n^* = \mathbf{OT}(\mathcal{T}_{e,n}, \mathcal{T}_b)$.
We define the FLOAT index of $\mathcal{T}_b$ in Eq.~\ref{eq:float_def} as $\lambda(\mathcal{T}_b)$.

\begin{align}
\lambda_n(\mathcal{T}_b)&=\sum\limits_{i=1}^{l_n}\sum\limits_{j=1}^{t_0}\mu_{n,i,j}^*c(o_{e,i}^n,o_{b,j}) \\
    \lambda(\mathcal{T}_b) &= \min\limits_{n} \lambda_n(\mathcal{T}_b)
    \label{eq:float_def}
\end{align}

Intuitively, FLOAT searches all expert demonstrations for the trajectory closest to $\mathcal{T}_b$ and uses the OT cost as the failure index.
If $\mathcal{T}_b$ stays far from all expert demonstrations under the OT metric, the rollout likely deviates from the task and fails.
Unlike prior methods that estimate policy uncertainty from single-step observations~\cite{chen2025uncertainty, xu2025can}, FLOAT relies on OT-based temporal alignment between rollout and expert trajectories, making it more robust to local errors and more aware of task progress.

For online failure detection, we derive a universal FLOAT threshold $\Lambda$ by calibrating on successful rollouts $\{\mathcal{T}_{b,k}\}_{k=1}^{M}$.
Specifically, $\Lambda$ is the $1-\delta$ percentile of $\{\lambda(\mathcal{T}_{b,k})\}_{k=1}^{M}$, where $\delta$ is a time-varying hyperparameter.
We raise a failure signal when the FLOAT index of a rollout exceeds $\Lambda$.
During rollout, $\delta$ is updated adaptively:
if the robot fails but FLOAT misses it, we decrease $\delta$ by $\Delta\delta$;
if FLOAT raises a failure signal while the robot is operating normally, we increase $\delta$ by $\Delta\delta$.
The threshold $\Lambda$ is then updated using $\delta$ and all successful rollouts gathered online. This adaptive update of $\delta$ helps prevent the FLOAT threshold from overfitting to anomalously high or low FLOAT indices.
FLOAT runs asynchronously with policy rollout to avoid latency from OT computation.

\subsection{ARMADA: Our multi-robot shared control system}
\label{subsec:multi-robot-sys}

With FLOAT, we can run autonomous policy rollouts in parallel across multiple robots.
To support this, we implement a message queue for communication between robot nodes and human teleoperation nodes.
Whenever FLOAT raises a failure signal, the robot node places a request for human intervention into the queue.
Teleoperation nodes pop the earliest request and assign it to an idle human operator, who then intervenes in the target rollout.
Algorithm~\ref{algo:multi-robot-sys} summarizes ARMADA.

\begin{figure*}[!t]
    \centering
    \includegraphics[width=0.8\linewidth]{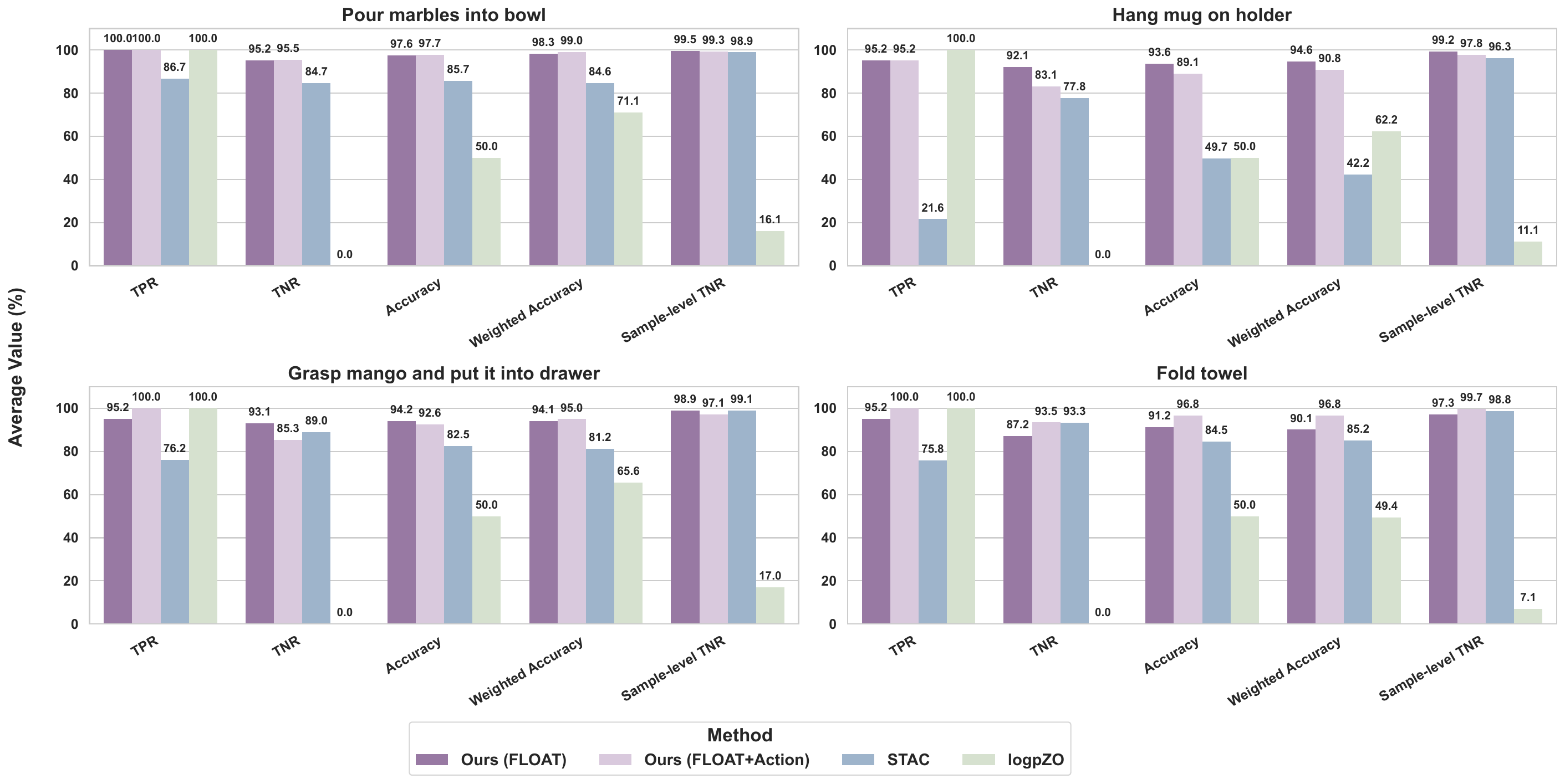}
    \vspace{0em}
    \caption{
    \textbf{Failure detection experiment results.} 
    FLOAT achieves nearly 95\% accuracy across four tasks, which is an improvement of over 20\% compared to state-of-the-art baseline methods. 
    It manifests comparable performance to its variant which further integrates action inconsistency metric, showcasing the effectiveness of FLOAT in detecting various failures.
    } 
    \label{fig:failure_detection_exp}
    \vspace{-3mm}
\end{figure*}

\begin{algorithm}[h]
    \scriptsize
    \caption{System design of ARMADA}
    \label{algo:multi-robot-sys}
    \DontPrintSemicolon 
    \SetKwBlock{DoParallelRobot}{do in parallel for all $R_i (1\leq i\leq m)$}{end}
    \SetKwBlock{DoParallel}{do in parallel}{end}
    \KwIn{Robot nodes $\{R_i\}_{i=1}^m$, Human teleoperation nodes $\{T_j\}_{j=1}^n$, Message queue $C$}
    \KwOut{Collected demonstration buffer $\mathcal{D}$}
    $\mathcal{D} \gets \emptyset$\;
    \DoParallelRobot{
        \While{True}{
            $R_i$ resets for new rollout episode \;
            \While{$R_i$'s episode not finished}{
                \eIf{detected failure}{
                    $C.put\_wait($``$R_i$ needs help''$)$\;
                }{
                Take one environment step\;
                }
            }
            $\mathcal{D}\gets \mathcal{D} \cup \{R_i$'s current episode$\}$ \;
        }
    }
    \DoParallel{
        \While{True}{
            \If{$C$ not empty}{
                $i\gets C.get\_robot\_idx()$; $C.pop()$\; 
                $j\gets Search\_for\_idle\_human\_node()$\;
                \DoParallel{$T_j$ takes over $R_i$ for failure recovery}
            }
        }
    }
    \Return{$\mathcal{D}$}\;
\end{algorithm}
\vspace{-4mm}

However, robots may still encounter unrecoverable states during rollouts.
For example, when a robot tries to grasp a cup of marbles but tips it over, the scattered contents make the task impossible to finish.
To handle such cases, we introduce an adaptive rewinding mechanism that returns the robot to an earlier timestep while human operators restore the scene.
This preserves intact, informative demonstrations enriched with human corrective behavior.
Moreover, the reset robot pose and scene configuration stay within the distribution of current policies.
This reduces the distribution shift between human teleoperators and learned policies, enabling efficient and stable policy learning.

Suppose a failure signal is raised at timestep $t_0$, and the current rollout trajectory is $(\mathcal{T}_b)_{1:t_0}=\{(o_{b,t}, a_{b,t})\}_{t=1}^{t_0}$.
We determine the rewind timestep using the objective in \ref{eq:rewind_obj}, where $\epsilon\in(0,1)$ is a predefined hyperparameter.
Intuitively, we find the latest timestep whose FLOAT index is below an adaptive threshold determined by the current FLOAT index.

\begin{equation}
    \argmax\limits_{t\geq 1} \mathds{1}[\lambda((\mathcal{T}_b)_{1:t})\leq \epsilon \lambda((\mathcal{T}_b)_{1:t_0})]\cdot t
    \label{eq:rewind_obj}
\end{equation}

This gives the robot a better chance to recover from failure and allows humans to help reset the scene.
We fix $\epsilon=0.2$ for all experiments in this work.

\subsection{Policy architecture and training recipe}
\label{subsec:policy_learning}

We use Diffusion Policy \cite{chi2023diffusion} as our imitation learning method and a transformer as the action-generation backbone.
We adopt a pretrained DINOv2 ViT-B/14 model \cite{oquab2023dinov2} as the visual encoder to obtain richer policy embeddings.
To compress the latent produced by DINOv2 for OT computation, we append a linear head to the visual encoder.

%% file: tex/5_experiment.tex
\section{Experiments}
\label{sec:exp}

We design the experiments to answer four questions:
(a) How does FLOAT perform compared to previous online failure detection methods?
(b) Does our rewinding mechanism help produce demonstrations with better quality?
(c) To what extent does ARMADA mitigate the reliance on human intervention over time?
(d) How does ARMADA perform in improving scalability of deployment and adaptation to unseen scenarios?

\begin{figure}[H]
    \vspace{-4mm}
    \centering
    \includegraphics[width=0.8\linewidth]{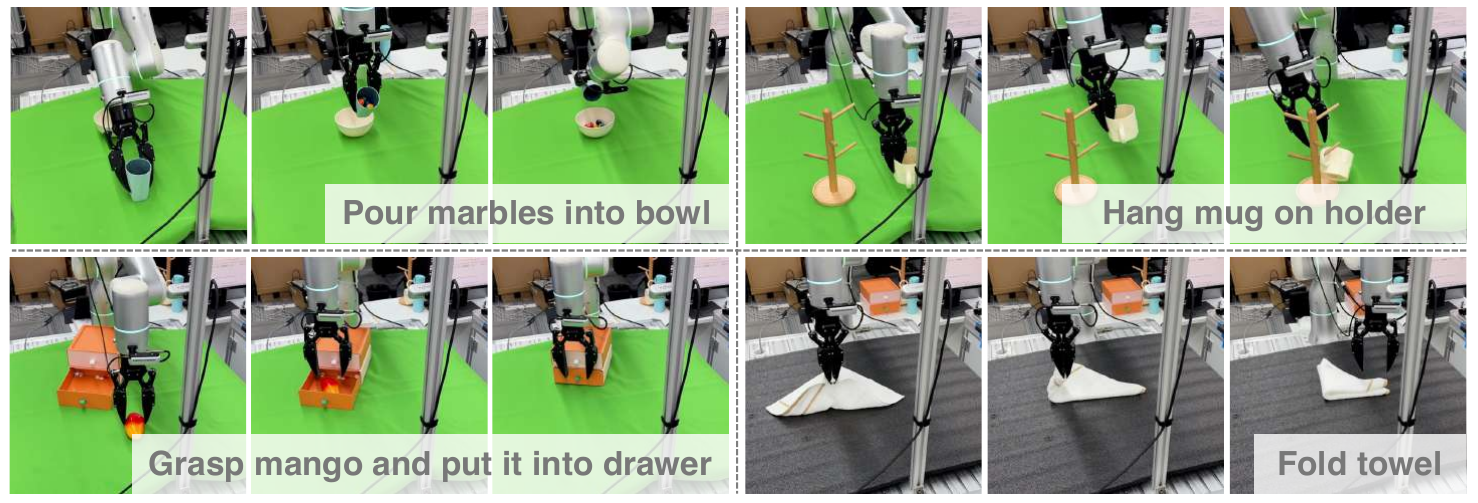}
    \caption{\textbf{Real-world task setup.}}
    \label{fig:task_setup}
    \vspace{-4mm}
\end{figure}

\textbf{Task design.} We adopt four real-world tasks: \textit{Pour marbles into bowl}, \textit{Hang mug on holder}, \textit{Grasp mango and put it into drawer}, and \textit{Fold towel}, illustrated in Figure \ref{fig:task_setup}.

\textbf{Hardware setup.} We use one eye-to-hand and one eye-in-hand Intel Realsense D435i camera for image observation.
Our robot platform is a Flexiv Rizon4 with a Robotiq 2F-85 gripper whose fingers are replaced by the UMI \cite{chi2024universal} gripper.
We use a Force Dimension sigma.7 haptic interface for teleoperation.

\textbf{Evaluation protocol.}
We initialize the training set with 50 human-teleoperated demonstrations per task.
In each rollout stage, we first run 30 episodes to calibrate the FLOAT threshold $\Lambda$, then freeze it and run 30 additional trials to evaluate failure detection while collecting online data for fine-tuning.
A failure is recorded when the robot does not complete the task within a task-specific time limit.
We use five failure-detection metrics: true positive rate (TPR), true negative rate (TNR), accuracy, weighted accuracy, and sample-level TNR.
A true positive is counted when the detector warns during a failed rollout; a true negative when it gives no warning during a successful rollout.
Accuracy and weighted accuracy are defined from TPR, TNR, and task success rate (SR), as shown in \ref{def:accuracy} and \ref{def:weighted_accuracy}.
We also introduce sample-level TNR, a step-level counterpart to the episode-level metrics, measuring true negative steps before policy failure.
This metric is important for multi-robot control systems, where frequent false positives would heavily burden throughput.

\begin{align}
    \text{Accuracy} &= \frac{\text{TPR}+\text{TNR}}{2} \label{def:accuracy}\\
    \text{Weighted Accuracy} &= \text{TPR} * \text{SR} + \text{TNR} * (1-\text{SR}) \label{def:weighted_accuracy}
\end{align}

After rollout, we perform offline fine-tuning by merging the collected online data with the initial human demonstrations to update the training buffer.
For each task, we run three rollout and two fine-tuning stages.
To reduce real-world variability, we strictly align the robot's initial pose and scene configuration for every trial within each rollout stage.

\subsection{How does FLOAT perform compared to previous online failure detection methods?}
\label{subsec:exp_fail_detect}

We compare FLOAT with two prominent online failure-detection baselines.
\textbf{STAC} \cite{agia2024unpacking}, the state-of-the-art method in our experiments, derives a time-invariant threshold by measuring the statistical distance between temporally overlapping regions of consecutive action chunk predictions.
\textbf{logpZO} \cite{xu2025can} trains a score model on policy embeddings from successful rollouts and calibrates time-varying thresholds with a Conformal Prediction (CP) band.

Figure~\ref{fig:failure_detection_exp} shows the failure-detection results.
FLOAT performs best, achieving nearly 95\% accuracy across four real-world tasks and improving over the state of the art by more than 20\%.
By contrast, logpZO suffers from severe overfitting and yields very low TNR.
We surmise this is because it assumes rollouts are Independently and Identically Distributed (I.I.D.), which does not hold in our experiments due to the random robot initial pose and large workspace.

\begin{figure}[htbp]
    \centering
    \includegraphics[width=0.49\linewidth]{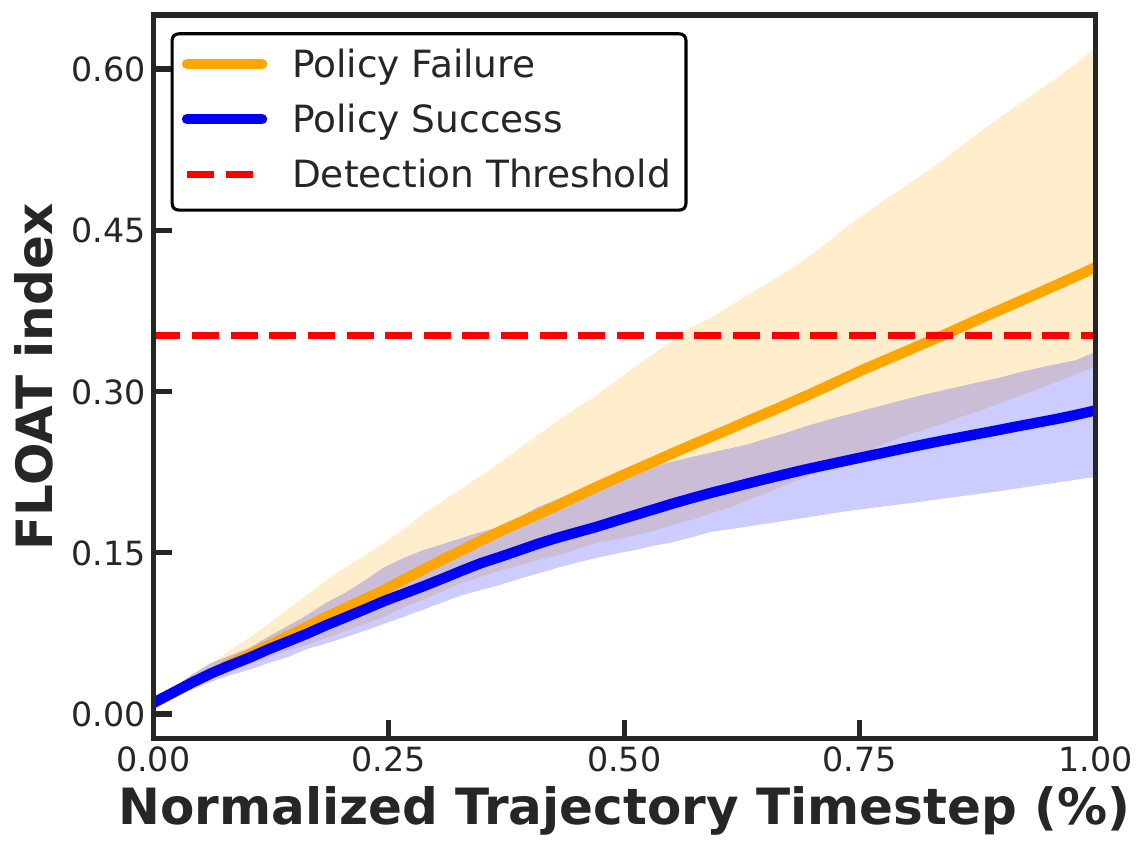}%
    \hfill
    \includegraphics[width=0.49\linewidth]{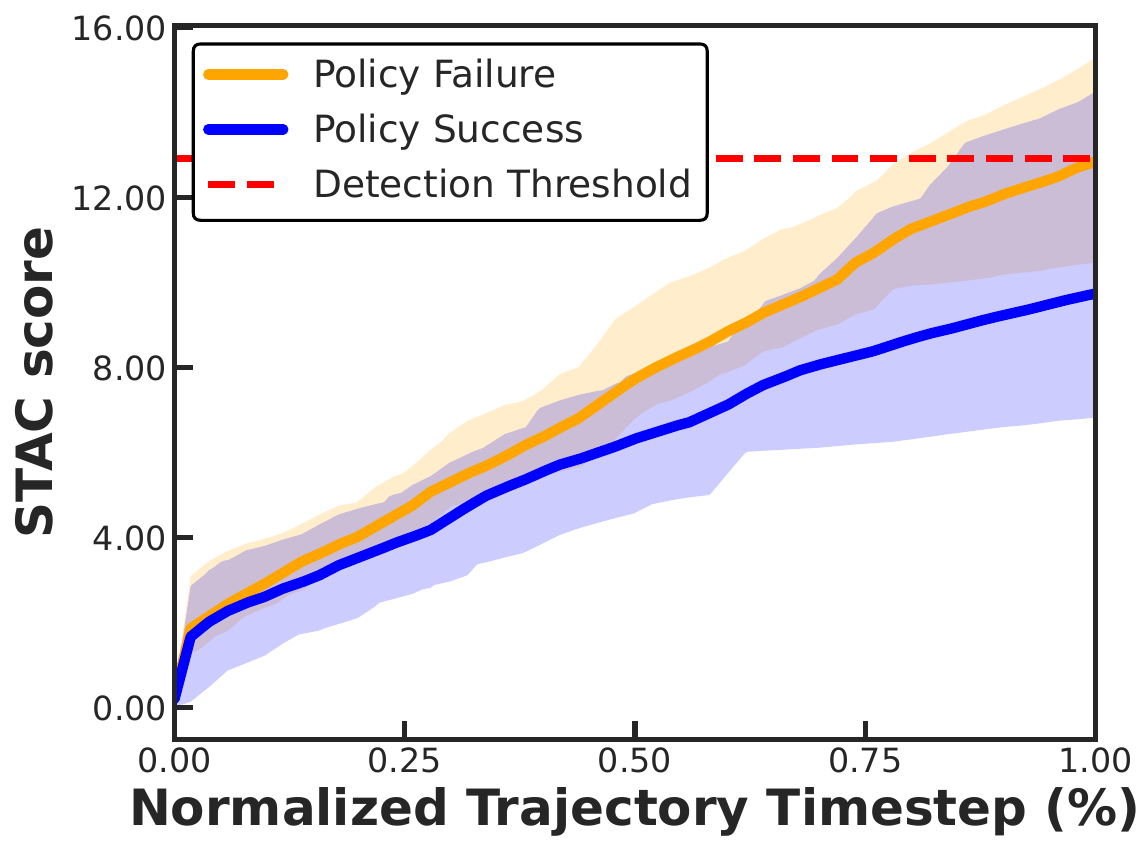}
    \caption{Quantitative results of FLOAT and STAC on \textit{Fold Towel}. Error bars indicate the 5-th and 95-th score quantiles.}
    \label{fig:qualitative_float}
    \vspace{-4mm}
\end{figure}

To further validate FLOAT, we introduce \textbf{FLOAT+Action}, a variant that combines FLOAT and STAC and raises a warning when either detector does.
FLOAT achieves comparable performance to this variant across all four tasks, showing that it can detect diverse failure modes without using action predictions.
We also present qualitative results for FLOAT and STAC on \textit{Fold Towel}. Their thresholds are pre-calibrated to the 95th percentile of successful rollouts during calibration. As shown in Figure~\ref{fig:qualitative_float}, FLOAT separates successful rollouts from failures more clearly and shows notably lower variance on successful rollouts.

\begin{table}[t]
\centering
\caption{Quantitative comparison with KNN and VIB-ILCtrl. FLOAT achieves the best (Weighted) Accuracy across both tasks, with substantially lower variance.}
\label{tab:all_results}
\scriptsize

\textbf{Hang mug on holder}
\vspace{1mm}

\resizebox{\columnwidth}{!}{%
\begin{tabular}{lccccc}
\toprule
\textbf{Method} & \textbf{TPR (\%)} & \textbf{TNR (\%)} & \textbf{Accuracy (\%)} & \textbf{Weighted Acc. (\%)} & \textbf{Step-level TNR (\%)} \\
\midrule
\textbf{FLOAT} &
77.96 {\scriptsize\textcolor{gray}{$\pm$ 7.99}} &
82.46 {\scriptsize\textcolor{gray}{$\pm$ 11.80}} &
\textbf{80.21} {\scriptsize\textcolor{gray}{$\pm$ 1.94}} &
\textbf{78.20} {\scriptsize\textcolor{gray}{$\pm$ 2.10}} &
96.80 {\scriptsize\textcolor{gray}{$\pm$ 3.60}} \\

\textbf{KNN} &
70.97 {\scriptsize\textcolor{gray}{$\pm$ 14.10}} &
32.06 {\scriptsize\textcolor{gray}{$\pm$ 17.10}} &
51.52 {\scriptsize\textcolor{gray}{$\pm$ 12.30}} &
51.93 {\scriptsize\textcolor{gray}{$\pm$ 17.80}} &
93.89 {\scriptsize\textcolor{gray}{$\pm$ 3.00}} \\

\textbf{VIB-ILCtrl} &
35.69 {\scriptsize\textcolor{gray}{$\pm$ 5.80}} &
91.51 {\scriptsize\textcolor{gray}{$\pm$ 1.17}} &
63.60 {\scriptsize\textcolor{gray}{$\pm$ 3.50}} &
62.76 {\scriptsize\textcolor{gray}{$\pm$ 9.20}} &
99.18 {\scriptsize\textcolor{gray}{$\pm$ 0.20}} \\
\bottomrule
\end{tabular}%
}

\vspace{2mm}

\textbf{Pour marbles into bowl}
\vspace{1mm}

\resizebox{\columnwidth}{!}{%
\begin{tabular}{lccccc}
\toprule
\textbf{Method} & \textbf{TPR (\%)} & \textbf{TNR (\%)} & \textbf{Accuracy (\%)} & \textbf{Weighted Acc. (\%)} & \textbf{Step-level TNR (\%)} \\
\midrule
\textbf{FLOAT} &
97.78 {\scriptsize\textcolor{gray}{$\pm$ 3.14}} &
91.26 {\scriptsize\textcolor{gray}{$\pm$ 4.40}} &
\textbf{94.52} {\scriptsize\textcolor{gray}{$\pm$ 2.30}} &
\textbf{95.71} {\scriptsize\textcolor{gray}{$\pm$ 2.30}} &
99.33 {\scriptsize\textcolor{gray}{$\pm$ 0.27}} \\

\textbf{KNN} &
86.94 {\scriptsize\textcolor{gray}{$\pm$ 10.90}} &
42.83 {\scriptsize\textcolor{gray}{$\pm$ 16.90}} &
64.89 {\scriptsize\textcolor{gray}{$\pm$ 4.20}} &
73.11 {\scriptsize\textcolor{gray}{$\pm$ 2.70}} &
93.33 {\scriptsize\textcolor{gray}{$\pm$ 3.50}} \\

\textbf{VIB-ILCtrl} &
42.78 {\scriptsize\textcolor{gray}{$\pm$ 25.60}} &
96.26 {\scriptsize\textcolor{gray}{$\pm$ 2.80}} &
69.52 {\scriptsize\textcolor{gray}{$\pm$ 13.00}} &
64.59 {\scriptsize\textcolor{gray}{$\pm$ 11.40}} &
99.81 {\scriptsize\textcolor{gray}{$\pm$ 0.14}} \\
\bottomrule
\end{tabular}%
}

\end{table}

\begin{figure}[t]
    \centering
    \includegraphics[width=0.8\linewidth]{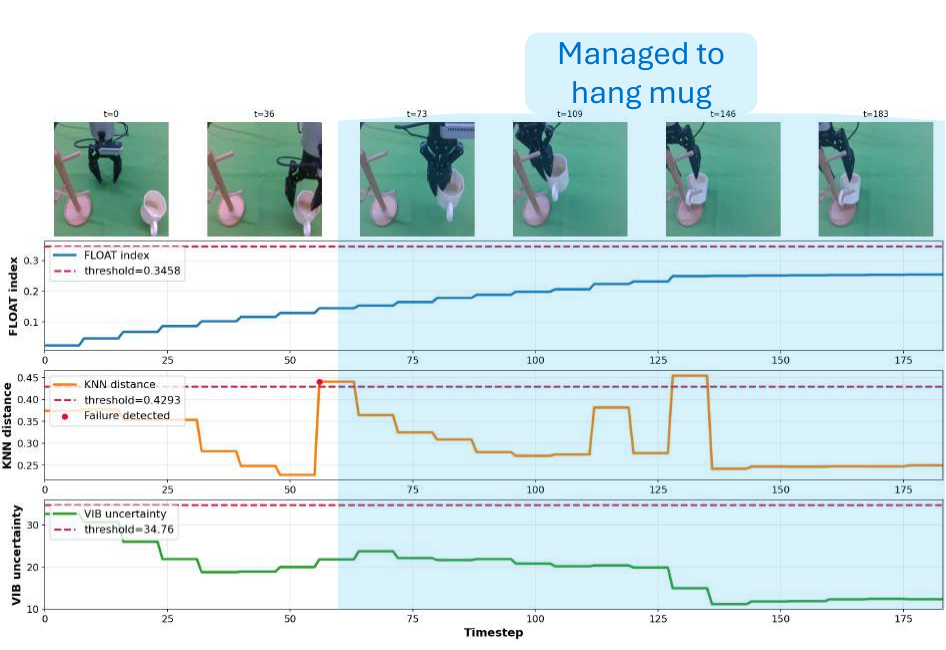}
    \vspace{-1mm}

    \includegraphics[width=0.8\linewidth]{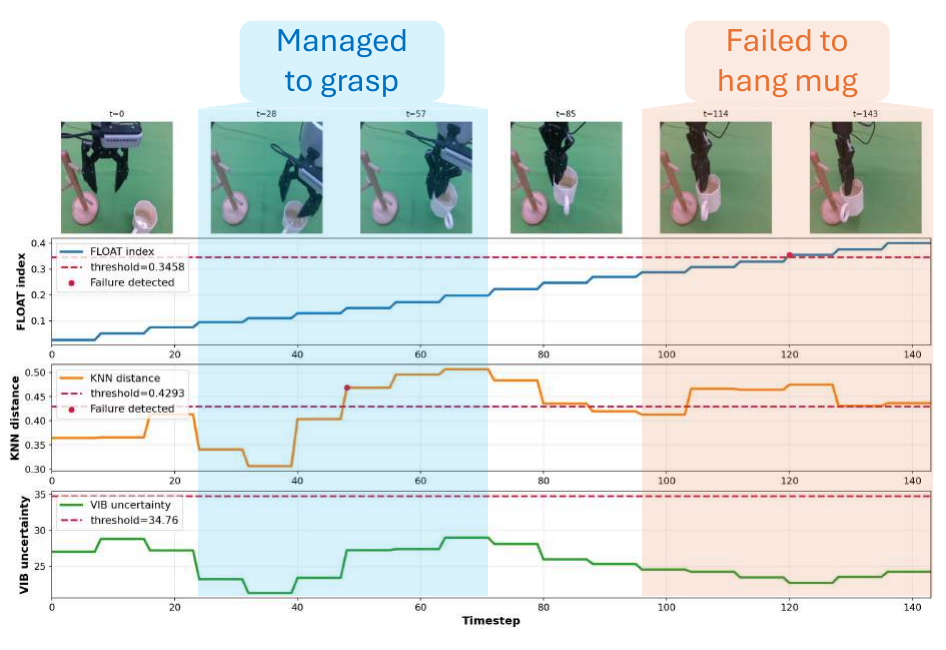}
    \caption{Qualitative comparison with KNN and VIB-ILCtrl. \textbf{Top: }The robot successfully hangs the mug on the holder. \textbf{Bottom: }The robot fails and drops the mug. FLOAT shows clearer temporal separation between successful and failed executions, leading to more accurate failure detection.}
    \label{fig:qualitative_knn_vib}
    \vspace{-4mm}
\end{figure}

To further validate the Optimal Transport formulation, we compare FLOAT with two additional baselines: \textbf{KNN}, which measures the average cosine distance between the current-state policy embedding and its $K{=}50$ nearest neighbours in the expert demonstrations, and \textbf{VIB-ILCtrl}~\cite{chen2025uncertainty}, which uses Variational Information Bottleneck (VIB) to compress the latent representation and estimate sample-wise uncertainty in latent space. As summarized in Table~\ref{tab:all_results}, FLOAT consistently outperforms KNN and VIB-ILCtrl in Accuracy and Weighted Accuracy on both \textit{Hang Mug} and \textit{Pour Marbles}, indicating a better balance between correctly identifying successful and failed executions. We attribute this advantage to the fact that KNN and VIB-ILCtrl detect failures from single-step embeddings, whereas FLOAT matches rollout trajectories and can better distinguish transient OOD states from sustained trajectory-level deviations. Figure~\ref{fig:qualitative_knn_vib} further shows clearer temporal separation between successful and failed executions for FLOAT on \textit{Hang mug}, supporting its advantage in accurate failure detection.

\begin{figure}[t]
    \centering
    \includegraphics[width=0.97\linewidth]{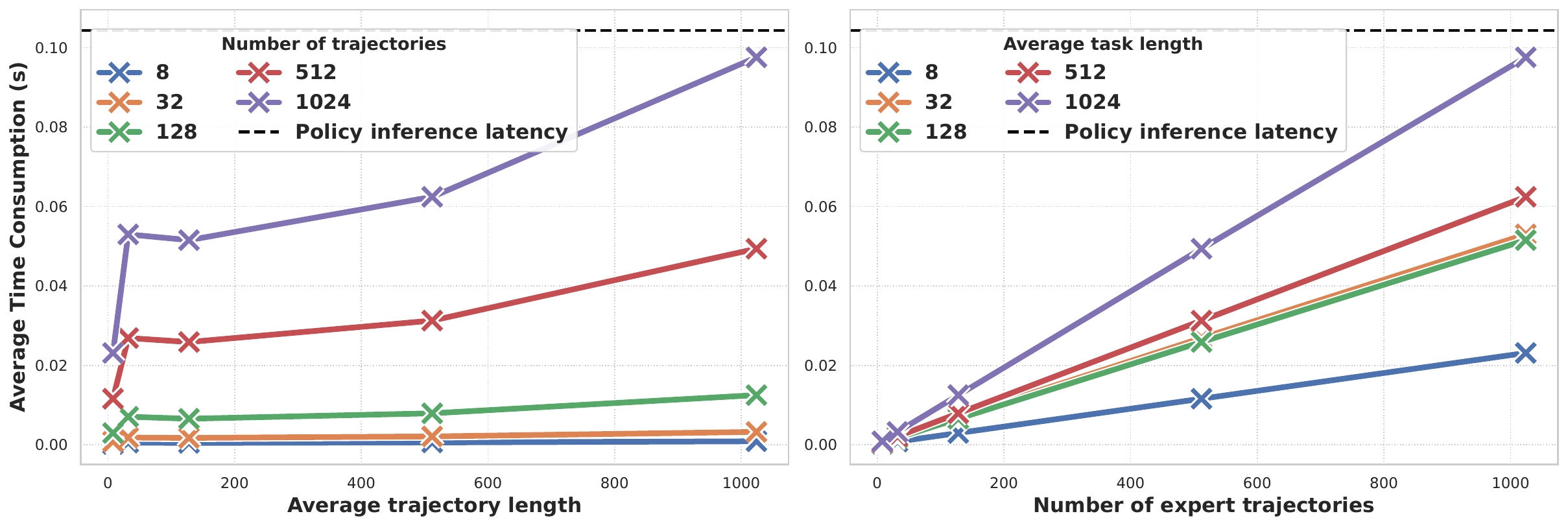}
    \caption{Runtime benchmark of FLOAT under varying numbers of expert trajectories and average task lengths.}
    \label{fig:float_time_cost}
    \vspace{-4mm}
\end{figure}

We further benchmark the runtime of FLOAT in Figure~\ref{fig:float_time_cost} under varying numbers of expert trajectories and average task lengths. FLOAT remains faster than policy inference even for task lengths around 100s and more than 1000 expert trajectories, indicating strong real-time capability. Runtime grows mainly with the number of trajectories and remains bounded when a fixed-size trajectory pool is used.

\subsection{Does our rewinding mechanism help produce demonstrations with better quality?}
\label{subsec:exp_sr_finetune}

\begin{figure}[ht]
    \centering
    \includegraphics[width=0.95\linewidth]{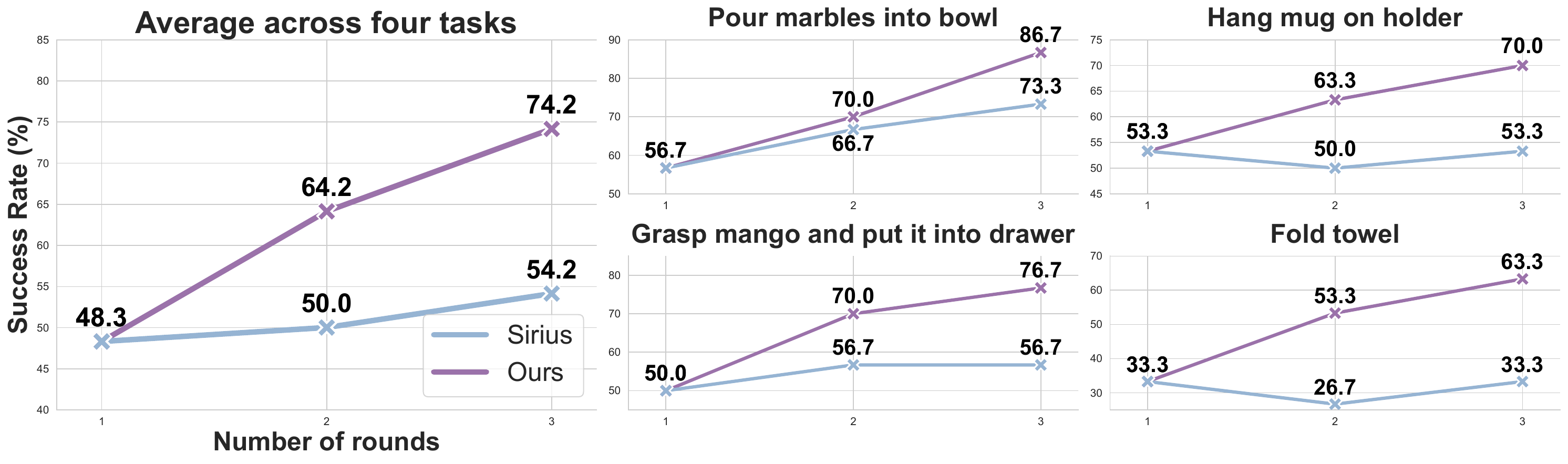}
    \vspace{0mm}
    \caption{\textbf{Success rate over three evaluation rounds.} ARMADA exhibits stable progress in success rate, with a more than four-fold increase compared to previous human-in-the-loop learning approach, thanks to our adaptive rewinding mechanism.}
    \label{fig:finetune_exp_sr}
    \vspace{-3mm}
\end{figure}

To demonstrate the effectiveness of adaptive rewinding, we compare policies trained on data collected by ARMADA and by the state-of-the-art human-in-the-loop method, \textbf{Sirius} \cite{liu2022robot}.
Sirius requires full-time human surveillance and reweights online observation-action samples for fine-tuning by emphasizing human intervention trajectories.
Figure~\ref{fig:finetune_exp_sr} shows the success-rate change for both methods.
Over three evaluation rounds, ARMADA improves success rate by 25.9\% on average, more than four times the improvement of Sirius.
In particular, ARMADA makes large gains on \textit{Hang mug on holder} and \textit{Fold towel}, where Sirius hardly helps.
This is due to the vulnerability of policies to unrecoverable states.
For example, if the robot fails to hang the mug on holder, the mug may fall onto the table and land on its side, making the task difficult even for human operators.

\begin{figure}[t]
    \centering
    \includegraphics[width=0.97\linewidth]{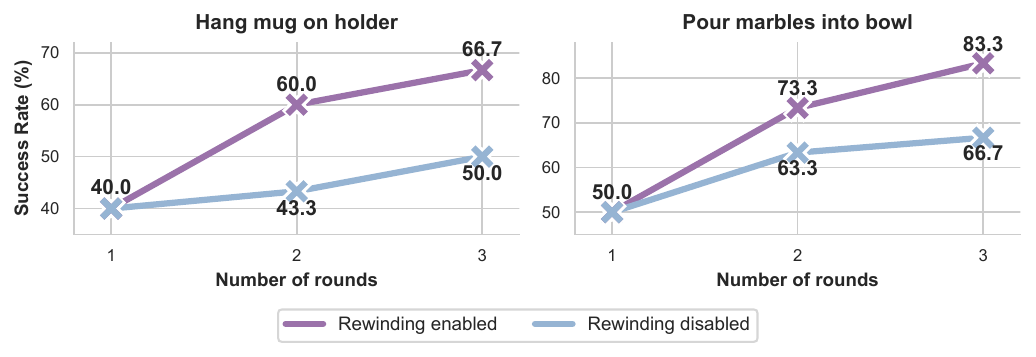}
    \caption{Ablation studies on rewinding mechanism. Comparing training with and without rewinding shows that rewinding substantially improves policy improvement by enabling recovery from hard failure states and yielding more useful intervention data.}
    \label{fig:rewinding_ablation}
    \vspace{-4mm}
\end{figure}

We further conduct an ablation study comparing policy improvement with and without rewinding. As shown in Figure~\ref{fig:rewinding_ablation}, removing rewinding leads to noticeably smaller gains, confirming its importance for recovering from unrecoverable states and collecting more informative interventions.

\subsection{To what extent does ARMADA mitigate the reliance on human intervention over time?}
\label{subsec:exp_human_ratio_finetune}

\begin{figure}[ht]
    \centering
    \includegraphics[width=0.95\linewidth]{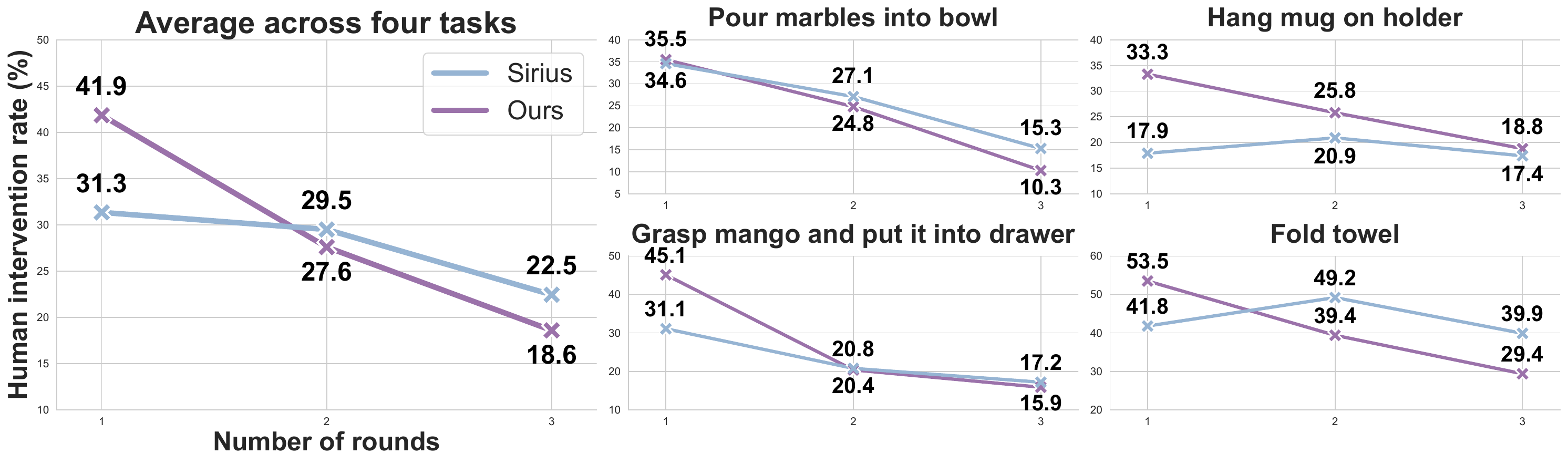}
    \vspace{0mm}
    \caption{\textbf{Human intervention rate over three evaluation rounds.} ARMADA results in a greater than two-fold reduction in human intervention rate compared to Sirius, showing potential in scalable deployment and adaptation.}
    \label{fig:finetune_exp_hr}
    \vspace{-4mm}
\end{figure}

In a human-in-the-loop learning framework, reliance on human intervention should decline as the policy improves through post-training.
To validate this, we inspect the human intervention rate of ARMADA and Sirius over three evaluation rounds.
As illustrated in Figure~\ref{fig:finetune_exp_hr}, although ARMADA requires more human intervention in the first evaluation round because of adaptive rewinding, it reduces the human intervention rate by an absolute 23.3\% after two fine-tuning stages. This reduction is more than twice that of Sirius, which achieves an 8.8\% reduction.
This suggests that ARMADA can improve real-world deployment scalability by increasing the number of paralleled robots in the multi-robot shared control system with comparable human effort.

\begin{figure}[ht]
    \vspace{0mm}
    \centering
    \includegraphics[width=0.8\linewidth]{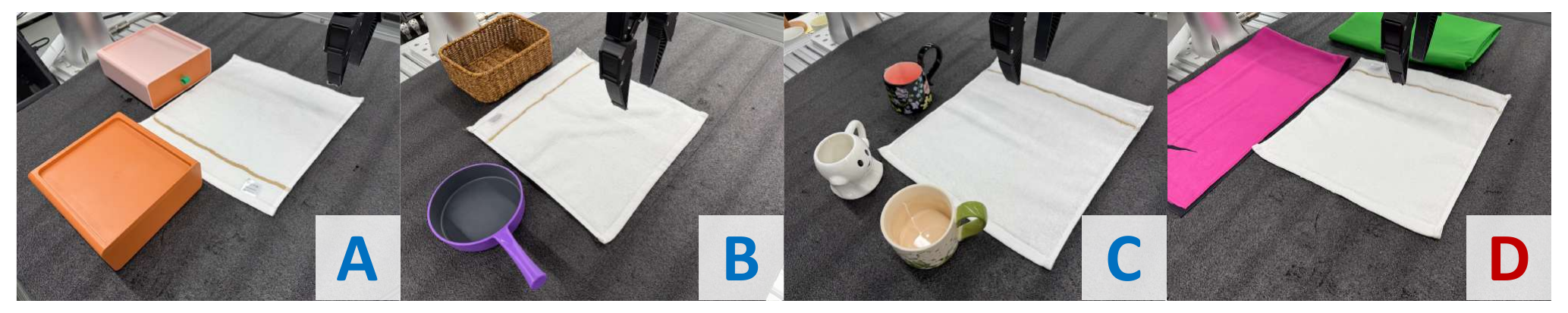}
    \caption{\textbf{Multi-robot experiments on unseen scenarios.} We deploy the pretrained \textit{Fold towel} policy on Scene A, B, and C \textcolor{myblue}{(in-domain)} for online data collection, and evaluate the post-trained policy on Scene D \textcolor{myred}{(out-of-distribution)}.}
    \label{fig:ood_exp}
    \vspace{-3mm}
\end{figure}

\subsection{How does ARMADA perform in improving scalability of deployment and adaptation to unseen scenarios?}
\label{subsec:exp_multi_robot_ood}

\begin{figure}[ht]
    \vspace{-3mm}
    \centering \includegraphics[width=0.7\linewidth]{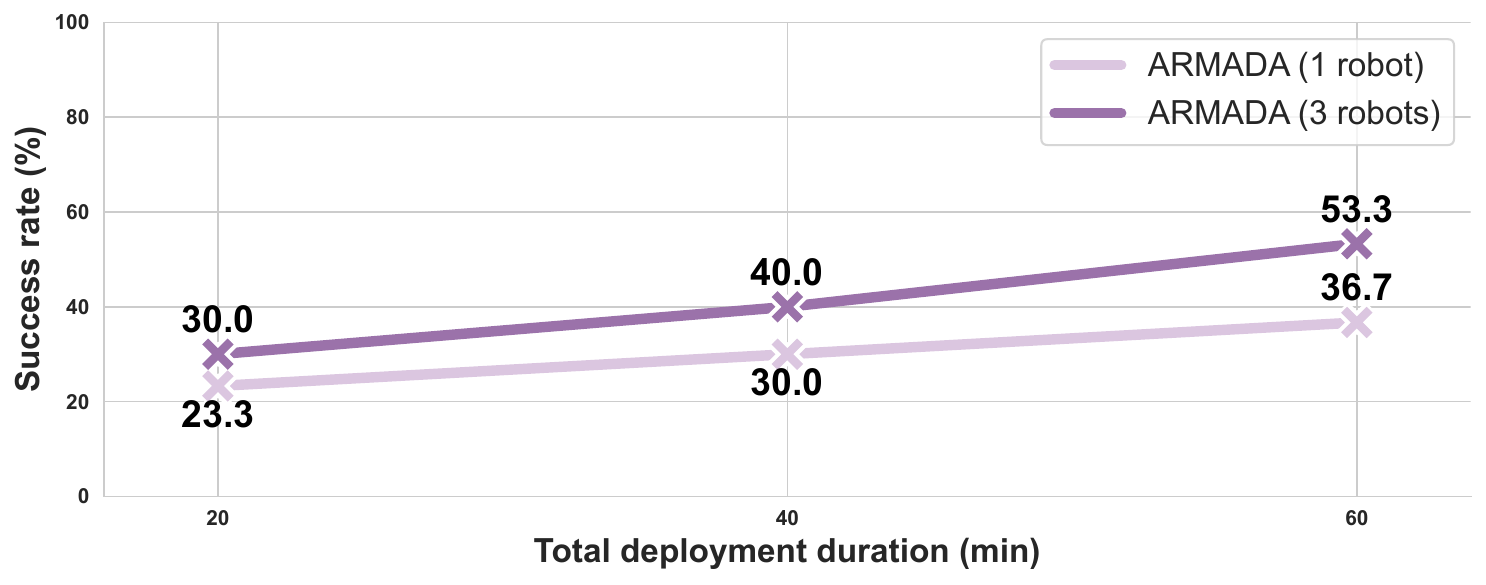}
    \caption{\textbf{Success rate on unseen scenarios.} ARMADA boosts adaptation to unseen scenarios with paralleled policy deployment on multiple robots.}
    \label{fig:ood_exp_success_rate}
    \vspace{-4mm}
\end{figure}

To verify ARMADA's effectiveness for adaptation to novel scenarios, we deploy a pretrained \textit{Fold towel} policy on three robots simultaneously for online data collection.
We also include three different sets of distractions during deployment, as shown in Figure \ref{fig:ood_exp}, to improve generalization of the post-trained policies to unseen domains.
As an ablation, we conduct another deployment round with the same pretrained policy only on Scene A in Figure \ref{fig:ood_exp}.
Each deployment round lasts 20 minutes and is followed by policy post-training.
Deployment and post-training alternate three times.
Each post-trained policy is evaluated on an unseen scenario, shown as Scene D in Figure \ref{fig:ood_exp}.
Figure~\ref{fig:ood_exp_success_rate} presents the success rates of all policies on the unseen scenario.

\begin{figure}[ht]
    \vspace{0mm}
    \centering \includegraphics[width=0.7\linewidth]{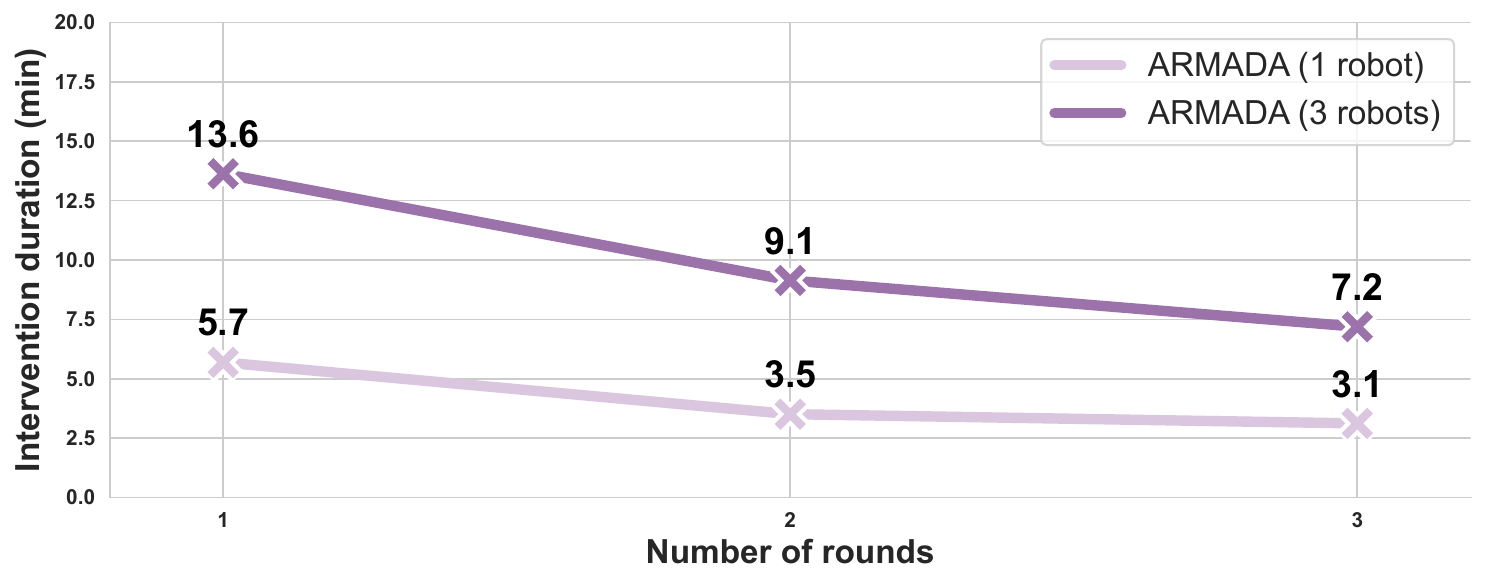}
    \caption{\textbf{Human intervention duration over three deployment rounds.} ARMADA scales up collection of human intervention trajectory with more robots in parallel.}
    \label{fig:ood_exp_human_time}
    \vspace{-3mm}
\end{figure}

Through parallel deployment on multiple robots and diverse scenes, ARMADA expedites adaptation to unseen scenarios and shows steady gains in task success rate as deployment time increases.
We attribute this to the growth of collected human intervention data and the increased diversity of scenarios as more robots are used.
To evaluate ARMADA's efficiency in collecting human intervention data from different domains, we also examine human-operator occupancy, measured by intervention duration in a single deployment stage.
Intuitively, a more occupied human operator indicates more efficient collection of valuable intervention data.
The results are shown in Figure~\ref{fig:ood_exp_human_time}.

For the same time budget, ARMADA benefits from parallel rollout and acquires more than twice as many human intervention samples as a single-robot setting.
These high-quality trajectories from diverse domains naturally improve policy robustness to novel scenarios, verifying ARMADA's practicality for more scalable policy adaptation and better use of human guidance.

\subsection{Failure Analysis}
\label{subsec:exp_failure_analysis}

\begin{figure}[t]
    \centering
    \phantomsection\label{fig:float_failure_analysis_false_positive}
    \includegraphics[width=0.86\linewidth]{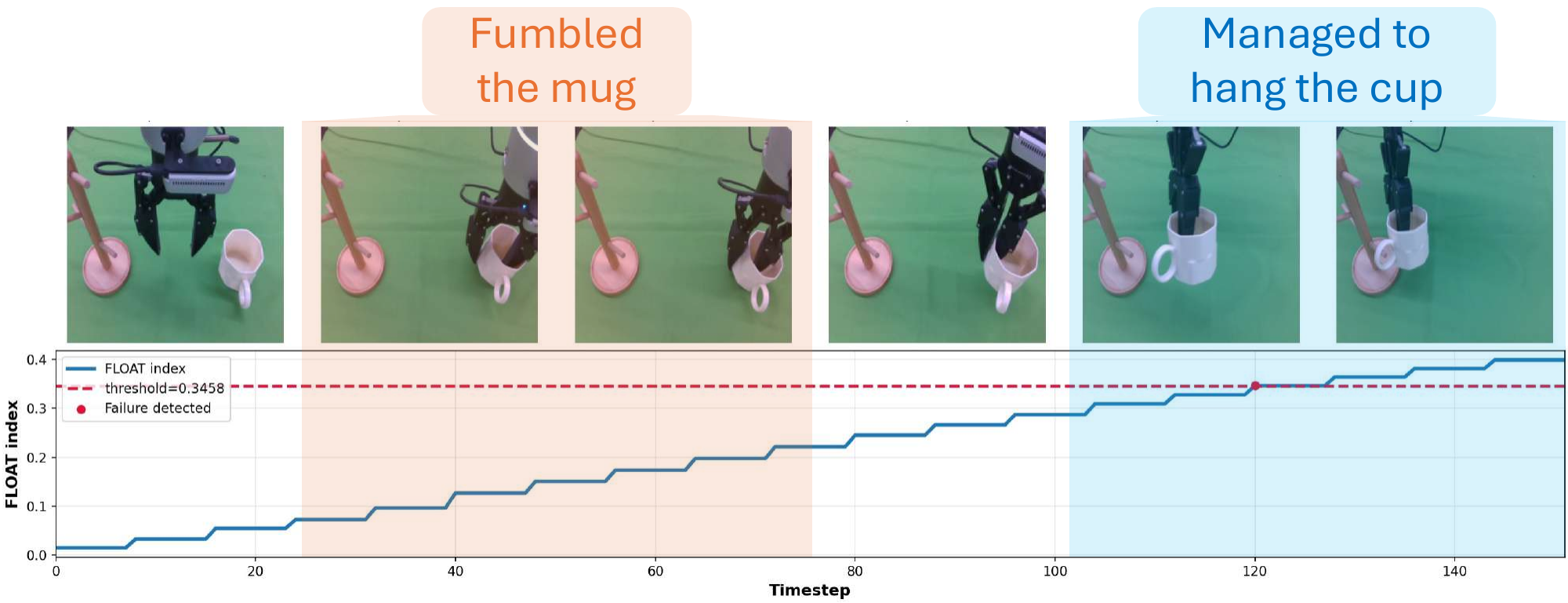}
    \vspace{2mm}

    \phantomsection\label{fig:float_failure_analysis_delayed}
    \includegraphics[width=0.86\linewidth]{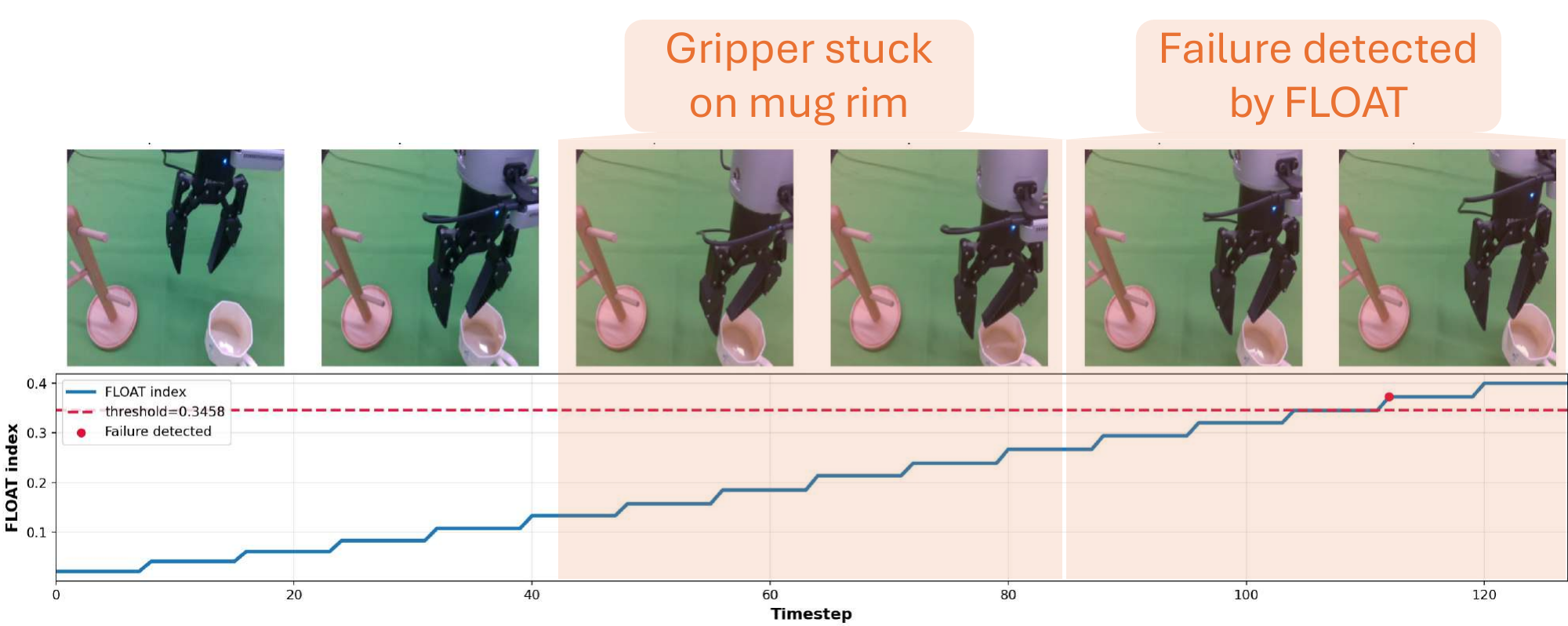}
    \caption{\textbf{Failure analysis of FLOAT.}
    (a) A false-positive sample where FLOAT raises a warning although the mug is eventually hung on the holder successfully.
    (b) A delayed-detection sample where FLOAT does not identify an early failure promptly and raises a warning after a noticeable delay.}
    \label{fig:float_failure_analysis}
    \vspace{-4mm}
\end{figure}

We analyze two representative failure cases of FLOAT in Figure~\ref{fig:float_failure_analysis}. In Figure~\ref{fig:float_failure_analysis_false_positive}, FLOAT produces a false positive when the robot spends unusually long trying to grasp the mug, causing the FLOAT index to rise even though the task eventually succeeds. In Figure~\ref{fig:float_failure_analysis_delayed}, FLOAT fails to detect an early failure promptly and raises a warning only after several additional timesteps. These examples suggest that FLOAT can both misclassify prolonged but recoverable executions and respond less promptly to subtle early failures.

\begin{figure}[t]
    \centering
    \includegraphics[width=0.85\linewidth]{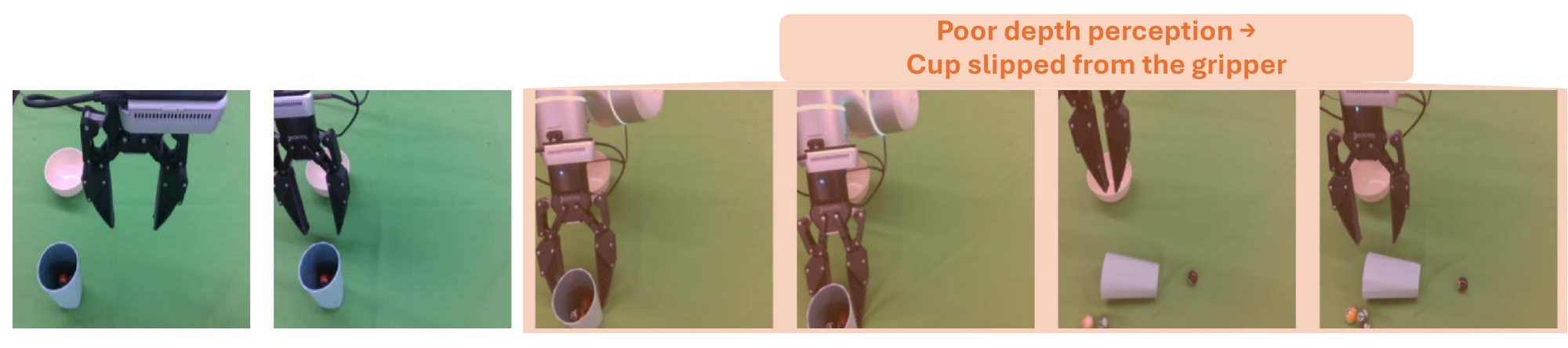}
    \vspace{-1mm}

    \includegraphics[width=0.85\linewidth]{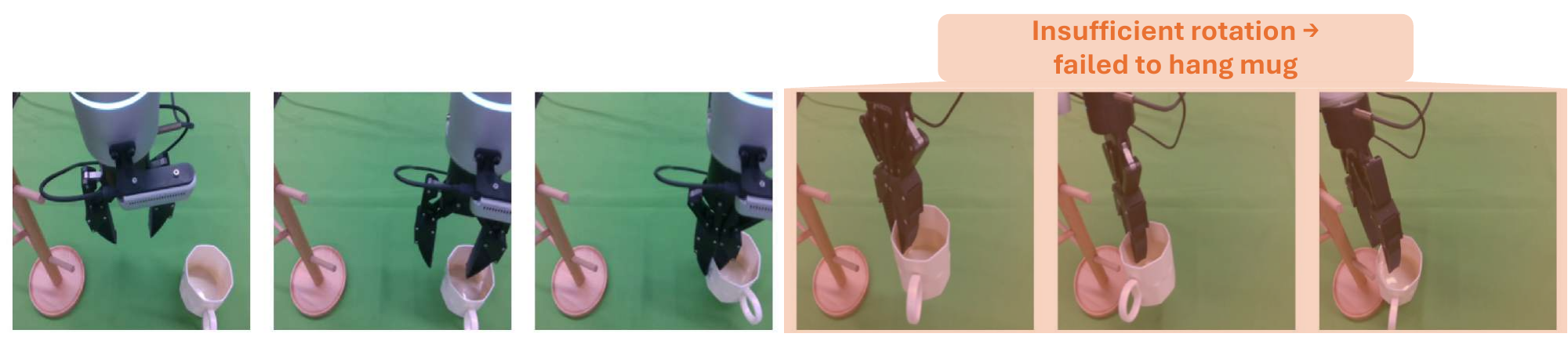}
    \caption{Representative policy failure cases after post-training with ARMADA. Top: \textit{Pour marbles into bowl}. Bottom: \textit{Hang mug on holder}. These cases mainly arise from perceptual ambiguity, such as inaccurate estimation of mug rotation or gripper grasp depth, rather than errors that can be fully corrected by DAgger-style behavioral supervision.}
    \label{fig:policy_failure_cases}
    \vspace{-4mm}
\end{figure}

We also examine representative failure cases that persist after post-training with ARMADA. As shown in Figure~\ref{fig:policy_failure_cases}, these failures are difficult for DAgger-style correction to fully resolve because they mainly stem from perceptual ambiguity rather than purely behavioral errors. In particular, the policy often fails to accurately perceive mug rotation or the gripper's grasp depth on the cup, leading to downstream execution failures. This suggests that such cases are fundamentally perception-limited and therefore cannot be completely corrected through behavioral supervision alone.

%% file: tex/6_discussion.tex
\section{Limitations and Discussion}

FLOAT still relies on expert demonstrations as references and thus does not naturally extend to novel tasks or embodiments. As discussed in Sec.~\ref{subsec:exp_failure_analysis}, it can also be slow to detect early failures. Since FLOAT is time-aware, unusually long task phases may cause false alarms, while subtle early failures may be detected late.
FLOAT also lacks action prediction, reacting to failures only after trajectory deviations appear rather than anticipating them.
Future work may develop more general progress estimators that transfer across tasks and hardware while incorporating predictive signals for earlier warnings, further improving scalable real-world deployment with systems such as ARMADA.
Furthermore, multi-robot deployment also requires human assistance for physical scene reset. While ARMADA reduces continuous human monitoring, reset-and-rewind still relies on humans to restore the environment. Thus, it should be viewed as a practical but temporary solution enabled by FLOAT, rather than a fundamental solution to predictive failure detection. Meanwhile, existing automatic scene reset methods are limited to reversible, relatively atomic manipulation skills such as pick-and-place. We therefore view scalable automatic scene reset and recovery as an important future direction, potentially combined with multi-robot systems like ARMADA toward fully autonomous deployment.

%% file: tex/7_conclusion.tex
\section{Conclusions}

This paper introduces ARMADA, a scalable multi-robot system for real-world deployment and adaptation, featuring FLOAT, an autonomous online failure detector that achieves nearly 95\% real-world accuracy.
By combining FLOAT with an adaptive rewinding mechanism, ARMADA significantly reduces human supervision over multiple post-training stages, substantially improving task success rate and reducing human intervention ratio.
However, FLOAT still has limitations, including limited predictive foresight and sensitivity to abnormal task-phase durations. Future work may address these issues with more predictive signals and more general progress estimators.